# Architecture inside the mirage: evaluating generative image models on architectural style, elements, and typologies

Jamie Magrill, MSc[1], Leah Gornstein[1], Sandra Seekins, MA[2], Barry Magrill, PhD[2*]

*Corresponding author barrymagrill@capilanou.ca

[1] *McGill University*
*845 Sherbrooke Street West*
*Montreal, Quebec, H3A 0G4, Canada*
[2] *Capilano University*
*2055 Purcell Way*
*North Vancouver, British Columbia, V7J 3H5, Canada*

January 13, 2026

## Abstract

Generative artificial intelligence (GenAI) text-to-image systems are increasingly used to generate architectural imagery, yet their capacity to reproduce accurate images in a historically rule-bound field remains poorly characterized. We evaluated five widely used GenAI image platforms (Adobe Firefly, DALL-E 3, Google Imagen 3, Microsoft Image Generator, and Midjourney) using 30 architectural prompts spanning styles, typologies, and codified elements. Each prompt–generator pair produced 4 images (n = 600 images total). Two architectural historians independently scored each image for accuracy against predefined criteria, resolving disagreements by consensus; set-level performance was summarized as 0–4 accurate images per four-image set. Image output from 'Common' prompts were 2.7-fold more accurate than 'Rare' prompts ($p < 0.05$). Across platforms, overall accuracy was limited (highest accuracy score 52%; lowest 32%, mean 42%), 'all-correct' (4/4) outcomes were similar across platforms; by contrast, 'all-incorrect' (0/4) outcomes varied substantially across platforms, with Imagen 3 exhibiting the fewest number of failures and Microsoft Image Generator exhibiting the highest number of failures. A qualitative review of our image dataset identified recurring patterns amongst GenAI images including over-embellishment, confusion between medieval styles and their later revivals, and misrepresentation of descriptive prompts (e.g., egg-and-dart, banded column, pendentive). These findings support the need for visible labels of GenAI synthetic content, provenance standards, particularly for future training datasets, and cautious educational use of GenAI architectural imagery in the pedagogical space.

# Introduction

Text-to-image generation systems, especially in latent diffusion models, have become both capable and commonplace. The latest text-to-image generation systems deliver striking improvements in visual plausibility and prompt adherence over the last half-decade[1-4], however, they are not without controversy: for instance, Google temporarily suspended Gemini's image generation of humans in February 2024 after widely reported outputs produced historically inaccurate depictions, and Google publicly pledged to relaunch the feature only after improving accuracy and safeguards[5,6]. Precisely because these tools generate outputs that emulate disciplinary conventions, they bring into focus a critical question for the fields of architecture and architectural history that is as evidentiary as it is aesthetic: do generative image models encode the style-specific knowledge that architecture and architectural history take for granted?

In architecture, the canon is image-born and based around sets of agreed-upon shared characteristics, rules, or physical attributes. Illustrated manuals, photographs, and pattern books including those found in textbooks, journal articles, and conference proceedings, have long standardized what counts as "Gothic," "Neoclassical," or "Byzantine," and have shaped everyday judgments about proportion, ornament, and tectonic logic[7-23]. The exponential increase in generative AI (GenAI) images (*i.e.,* synthetic content) found online, many of which could eventually become indexed by standard search engines, risks a silent distortion of the shared visual commons that supports architectural and architectural history pedagogy.

Major GenAI platforms and infrastructure providers are rolling out provenance tools to address these concerns, such as Content Credentials (C2PA) to embed creation-time metadata that persist after certain forms of online distribution, SynthID watermarking, and platform-level labels affixed to GenAI images (**Figure 1**). Governments are acting to address concerns around GenAI images as well: the 2024 EU AI Act (*Artificial Intelligence Act (Regulation (EU) 2024/1689)*)[24] requires that synthetic media (e.g., deepfakes) be marked as artificially generated, and in the U.S. the 2023 Executive Order on AI (*Executive Order 14110 on Safe, Secure, and Trustworthy Development and Use of Artificial Intelligence*)[25] directs agencies to advance watermarking and content-authentication guidance, following synthetic-content guidance from the U.S. AI Safety Institute[26-30]. In practice, however, the rollout of durable markings (visible labels, embedded metadata like C2PA, and/or hard-to-remove watermarks) of GenAI images has been inconsistent, with a mixture of visible and non-visible labels, coupled with a patchwork of directives from governments and international regulatory agencies over the past 2-3 years, leading to a minority of GenAI text-to-image systems implementing watermarks and deepfake labelling[30].

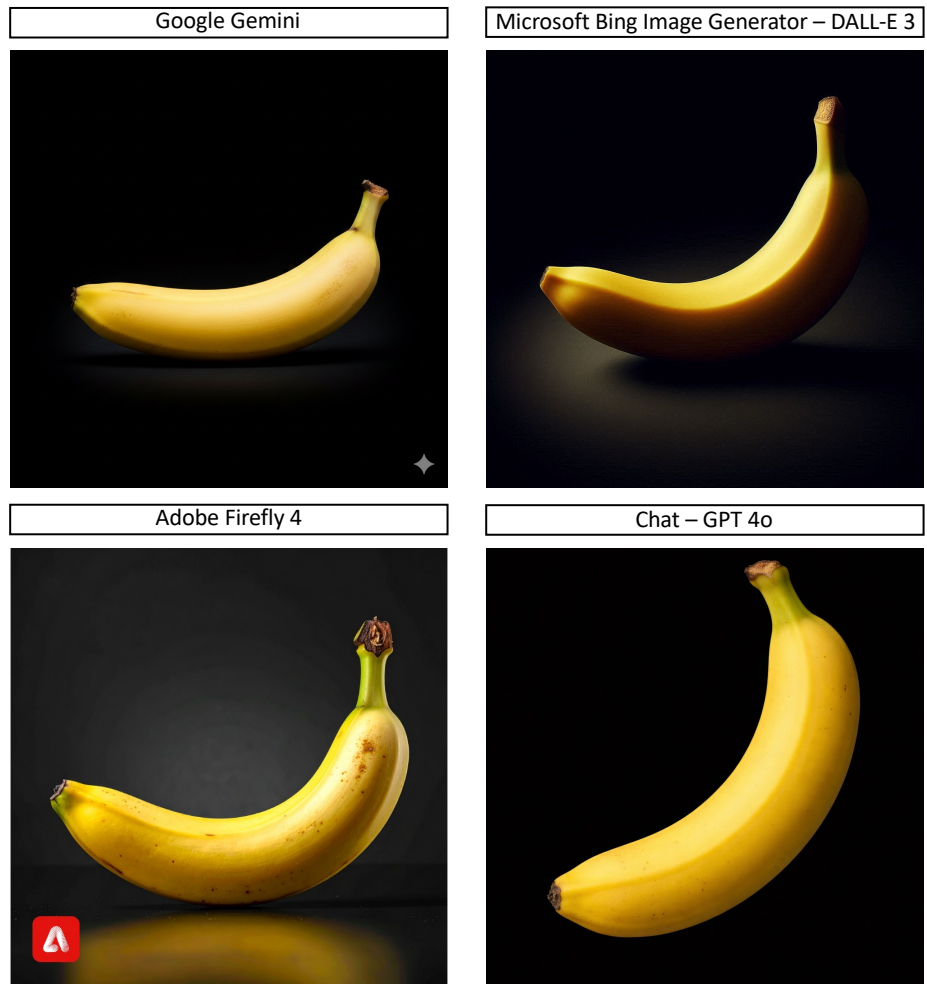

**Figure 1. Examples of generative-AI watermarks and platform labeling conventions.** Representative images generated from an identical simple prompt: "Create an image of a banana on a black background" illustrate how provenance signals and visible watermarks are currently implemented inconsistently across platforms. Representative images are provided for illustration only and were not part of the image generation dataset.

Two additional concepts require a careful, style-aware evaluation in this paper. First, contemporary vision systems tend to privilege textural cues over global shape and proportion; that is, GenAI models primarily recognize objects by textures rather than shapes by focusing on local, repetitive patterns of pixels rather than the global, structural outline of an object, unlike humans who have a strong shape-recognition bias[31-33]. Second, if future GenAI models are trained on datasets saturated with synthetic images, the possibility exists that future image quality could degrade via "model collapse," which creates some urgency to ensure appropriate dataset curation and image provenance for training datasets going forward[34-36].

We therefore treat architectural imagery as a *stress test* for whether generative systems can faithfully reproduce rule-bound stylistic conventions. We examine model fidelity across a variety of styles, forms and other qualitative groupings using a 600-image dataset. We also quantify generator-level differences in "all-correct" and "all-incorrect" outcomes, with implications for classroom use, search, and design workflows. Our results show systematic over-detailing and over-embellishment of ornamentation, motifs and design features, confusion between earlier-era styles (e.g., medieval, renaissance) and their 19th-century revivals (e.g., gothic revival, neo-clacissism), and recurrent misinterpretations of codified motifs (e.g., egg and dart, pendentive, Greek key).

## Methods

*Study Design and Data Collection*

We evaluated five GenAI image generators: Adobe Firefly, DALL-E 3, Google Imagen 3, Microsoft Image Generator, and Midjourney on a fixed list of 30 prompts spanning a diverse collection of styles, forms, eras, typologies/motifs, and classical orders. Image generation was undertaken in January 2025. For each prompt–generator pair we collected four images (a "set"), yielding n = 120 prompt–generator pairs using the following open-ended stems: "Create an image of a building in the style of [architectural style]" or "Create an image of a [architectural motif]" (**Table 1**). All prompt–generator pairs were successful except for 'Pendentive' for Adobe Firefly resulting in n = 600 images including n = 4 failed images (see **Extended Data 1**). All images in our image generation set may be found in **Extended Data 1**. Image generation used default settings; seeds were left unset to preserve platform typical stochasticity. Midjourney's /imagine grid was treated as a four-image set; platforms returning one image per request were sampled four times.

*Image Evaluation and Scoring*

Two architectural historians (SS, BM) independently scored each image as accurate (1) or inaccurate (0) for the requisite style or motif, emphasizing: (i) proportional and ornamental correctness; (ii) structural plausibility; and (iii) historical authenticity and referring to a pre-set list of criteria for all edge-cases (**Table 1**). Disagreements were resolved to a final consensus at the image level through discussion. A careful visual review of all 600 images was conducted by one architectural historian (BM) and one additional researcher (JM) to analyze patterns in image-sets. Image set-level scores equal the sum of accurate images per set (0–4). Figures display representative accurate/inaccurate pairs for exemplar prompts. Categories and groupings for analysis were assigned by BM.

*Outcomes*

The primary outcome of this paper was the image set-level accuracy score (0–4) for each prompt. Secondary outcome included 'all-correct' (4/4) and 'all-incorrect' (0/4) set counts.

*Statistical Analyses*

For tier-based comparisons we utilized per-style means (each style contributes its mean image-set score across all 5 GenAI image generators). Error bars were calculated as Standard Error-of-Mean. Differences in pairwise categorical variables were assessed with Mann–Whitney tests. Group differences in categorical variables were assessed using the Kruskal–Wallis test with Dunn's post-hoc tests for pairwise comparisons, with p-values adjusted for multiple testing where appropriate. Pairwise differences in set-level accuracy between generators were assessed using paired t-tests. Differences in the proportion of 'all-correct' (4/4) and 'all-incorrect' (0/4) image sets across rarity tiers were evaluated using a chi-square test of independence, and an ordered chi-square test for trend (Cochran–Armitage) was performed to assess monotonic changes across tiers. Pairwise comparisons between tiers were conducted using two-sided Fisher's exact tests, with Holm correction applied to control the family-wise error rate. We performed all analyses using GraphPad Prism version 10.4.0 (GraphPad Software, San Diego, CA, USA).

# Results

*Quantitative Results*

We first divided our set of 30 prompts based on general prevalence in the architectural history lexicon into three rarity tiers: Common, Intermediate and Rare (**Figure 2A**). This tiering surfaced two broad patterns. First, across all three tiers (Common, Intermediate and Rare), we observed a wide distribution and range of accuracy, with some prompts (e.g., High Gothic, Common; Neo-Classicism, Intermediate; Egyptian Revival, Rare) generating images with a higher degree of accuracy, while other prompts (e.g., French Rococo, Common; Carolingian, Intermediate; Banded Column, Rare) generated images with a much lower degree of accuracy. Second, we observed an overall trend towards lower per-style means with increasing rarity, reaching significance for Common vs. Rare styles (Means - Common: 2.05, n=15, Rare: 0.76, n=5, p=0.049) (**Figure 2B**).

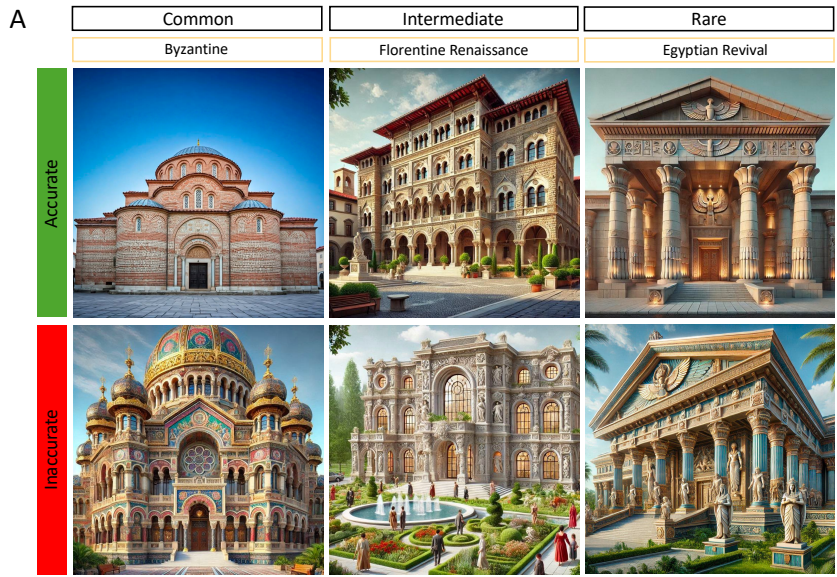

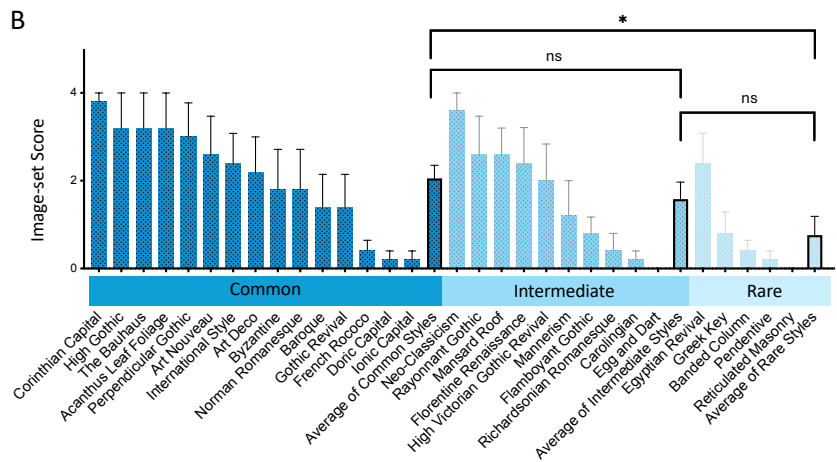

**Figure 2. Styles by rarity tier: representative accurate and inaccurate outputs and tier-level performance.** (A) Exemplary "accurate" (top row) and "inaccurate" (bottom row) outputs for three prompts selected to represent each rarity tier used in the analysis (Common: *Byzantine*; Intermediate: *Florentine Renaissance*; Rare: *Egyptian Revival*). (B) Summary comparison of set-level accuracy across rarity tiers. Each prompt–platform pair produced a four-image set (0–4 accurate images), and per-style means were calculated across all five GenAI image generators. Error bars were calculated as Standard Error-of-Mean. Differences in pairwise categorical variables were assessed with Mann–Whitney tests, significance was reached at p < 0.05.

We next organized our set of 30 prompts by Era (Medieval, Renaissance, Revival and Modern) and Style (Medieval-Renaissance and Revival) (**Figure 3A**). Here, we observed an overall trend towards lower per-style means for Styles as opposed to Eras (Means - Eras: 1.95, n=20, Style: 1.14, n=10, p=0.079) though not for any individual groupings (**Figure 3B**).

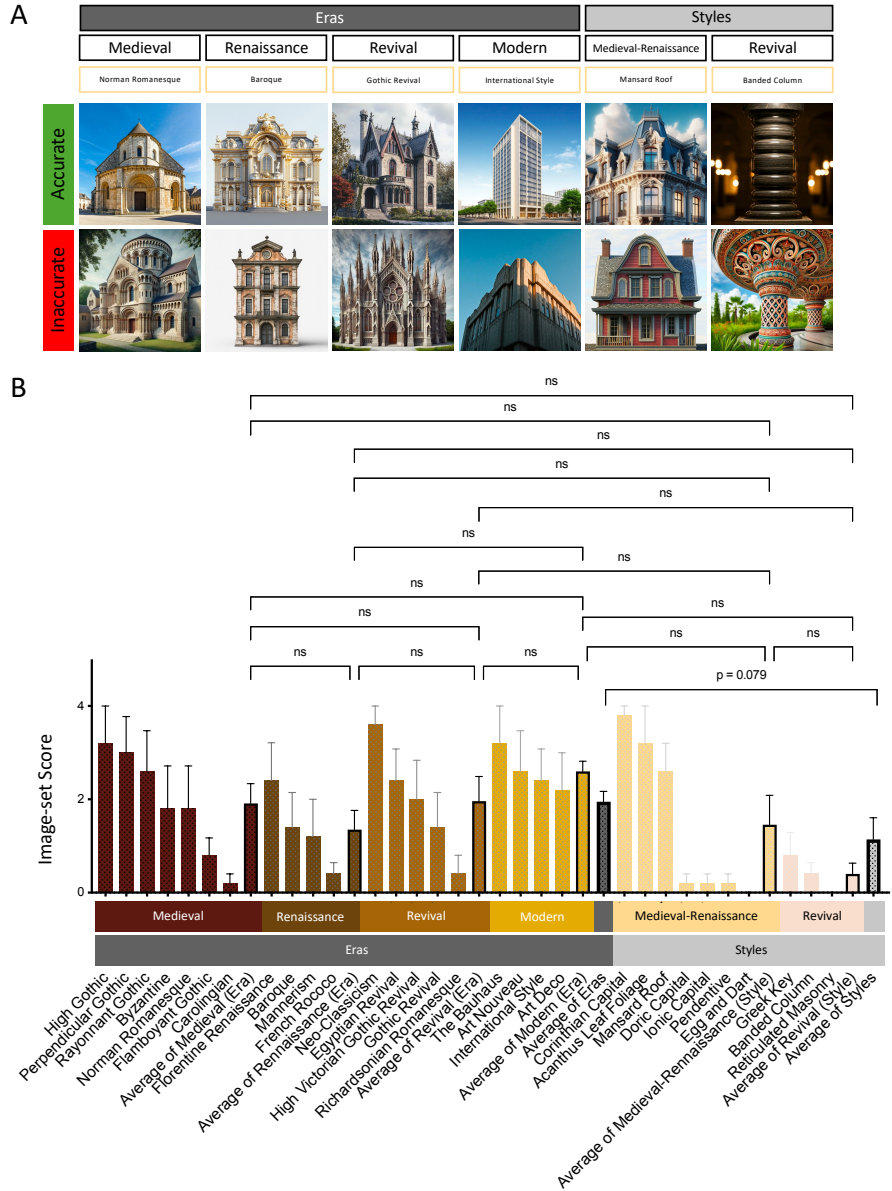

**Figure 3. Styles and eras: representative accurate and inaccurate outputs and aggregate performance by historical grouping.** (A) Exemplary "accurate" (top row) and "inaccurate" (bottom row) outputs for six prompts selected to represent each rarity tier used in the analysis grouped by Eras (Medieval, Renaissance, Revival, Modern) and by Styles (Medieval–Renaissance, Revival). (B) Summary comparison of set-level accuracy across historical groupings and aggregate comparison of mean set-level accuracy between prompts categorized as "Eras" versus "Styles". Each prompt–platform pair produced a four-image set (0–4 accurate images), and per-style means were calculated across all five GenAI image generators. Error bars

were calculated as Standard Error-of-Mean. Differences in pairwise categorical variables were assessed with Mann–Whitney tests, group differences in categorical variables were assessed using the Kruskal–Wallis test with Dunn's post-hoc tests for pairwise comparisons, with p-values adjusted for multiple testing where appropriate, significance was reached at p < 0.05.

Finally, we organized our set of 30 prompts by style and formal tendency (Structuralist, Post-Structuralist, Plasticizing, Historicist, Geometric and Foliated (**Figure 4A**). Overall, we observed no significant differences between groups, although we did observe a trend between Plasticizing (a form emphasizing a prominent three-dimensional, sculptural quality) and Geometric (an architectural style employing shapes, patterns and proportions) styles (Means - Geometric: 2.6, n=3, Plasticizing: 0.28, n=5, p=0.15) (**Figure 4B**).

**Figure 4. Styles and eras: representative accurate and inaccurate outputs and aggregate performance by stylistic and formal tendency.** (A) Exemplary "accurate" (top row) and "inaccurate" (bottom row) outputs

for six prompts selected to represent each rarity tier used in the analysis grouped by stylistic and formal tendency (Structuralist, Post-Structuralist, Plasticizing, Historicist, Geometric, Foliated). (B) Summary comparison of set-level accuracy across stylistic and formal tendency groupings. Each prompt–platform pair produced a four-image set (0–4 accurate images), and per-style means were calculated across all five GenAI image generators. Error bars were calculated as Standard Error-of-Mean. Differences in pairwise categorical variables were assessed with Mann–Whitney tests, group differences in categorical variables were assessed using the Kruskal–Wallis test with Dunn's post-hoc tests for pairwise comparisons, with p-values adjusted for multiple testing where appropriate, significance was reached at p < 0.05.

We then investigated overall accuracy across all five GenAI text-to-image systems (Adobe Firefly, DALL-E 3, Google Imagen 3, Microsoft Image Generator, and Midjourney), ordered by commonality, era, style, and form (**Figure 5A**). We observed a range of accuracy scores across GenAI text-to-image systems; the highest accuracy score of any GenAI text-to-image system was 52% (Imagen 3), while the lowest was 32% (Microsoft Image Generator), with an overall cross-platform average of 42% (**Figure 5B**). We observed no significant differences in overall accuracy between any of the five GenAI text-to-image systems.

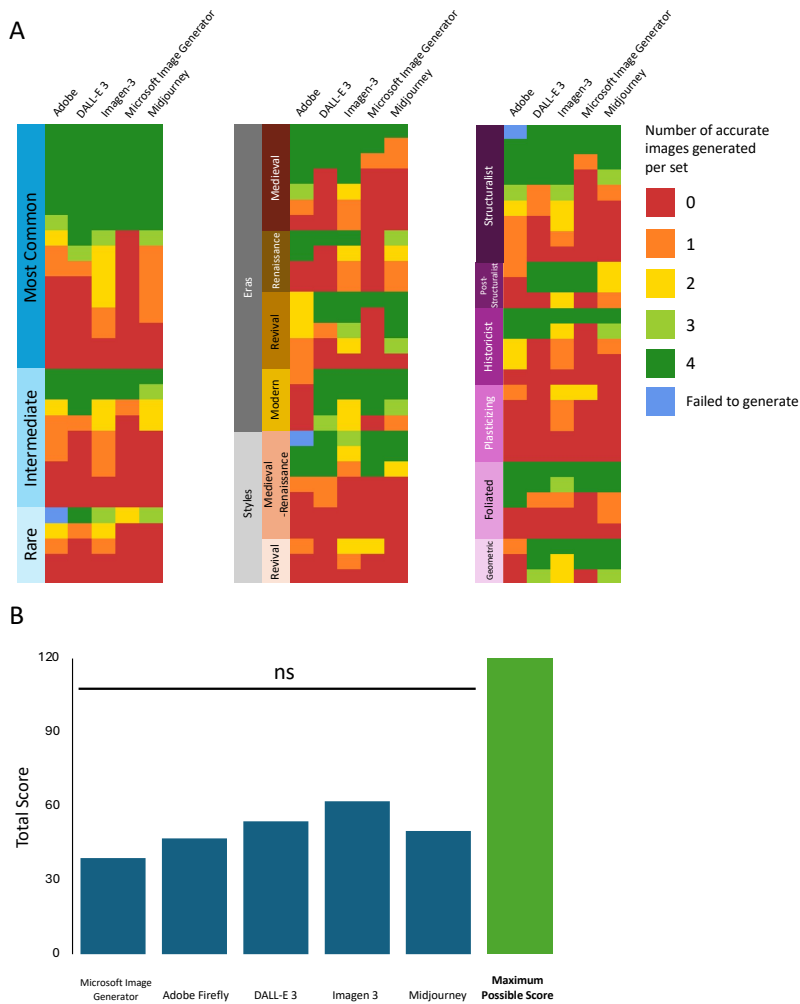

**Figure 5. Heatmaps of set-level accuracy across GenAI text-to-image generators.** Heatmaps summarize set-level performance for each platform across the 30 prompts. Each cell encodes the number of accurate images in a four-image set (0–4), with "failed to generate" denoted separately. Prompts are organized into three complementary orderings used throughout the analysis: by rarity tier, by historical era and style grouping, and by stylistic and formal tendency groupings. (B) Total accuracy score per GenAI text-to-image generator (maximum possible score of 120 per generator). Pairwise differences in set-level accuracy between generators were assessed using paired t-tests, significance was reached at p < 0.05.

We next wanted to explore the frequency of 'all-correct' (4/4) and 'all-incorrect' (0/4) outcomes. We observed a narrow range of 'all-correct' (4/4) outcomes across all five GenAI text-to-image systems with a correlation between rarity and 'all-correct' outcomes The highest frequency of 'all-correct' outcomes was DALL-E 3 (40%), while the lowest frequencies were Adobe Firefly and Midjourney (27%), though we did not observe significant differences in 'all-correct' (4/4) outcomes between any of the five GenAI text-to-image systems (**Figure 6A-B**). We observed a wider range of 'all-incorrect' (0/4) outcomes across all five GenAI text-to-image systems; the highest frequency was Microsoft Image Generator (63%), while the lowest frequency was Imagen-3 (20%) (p < 0.01); no other comparisons reached statistical significance (**Figure 6C-D**).

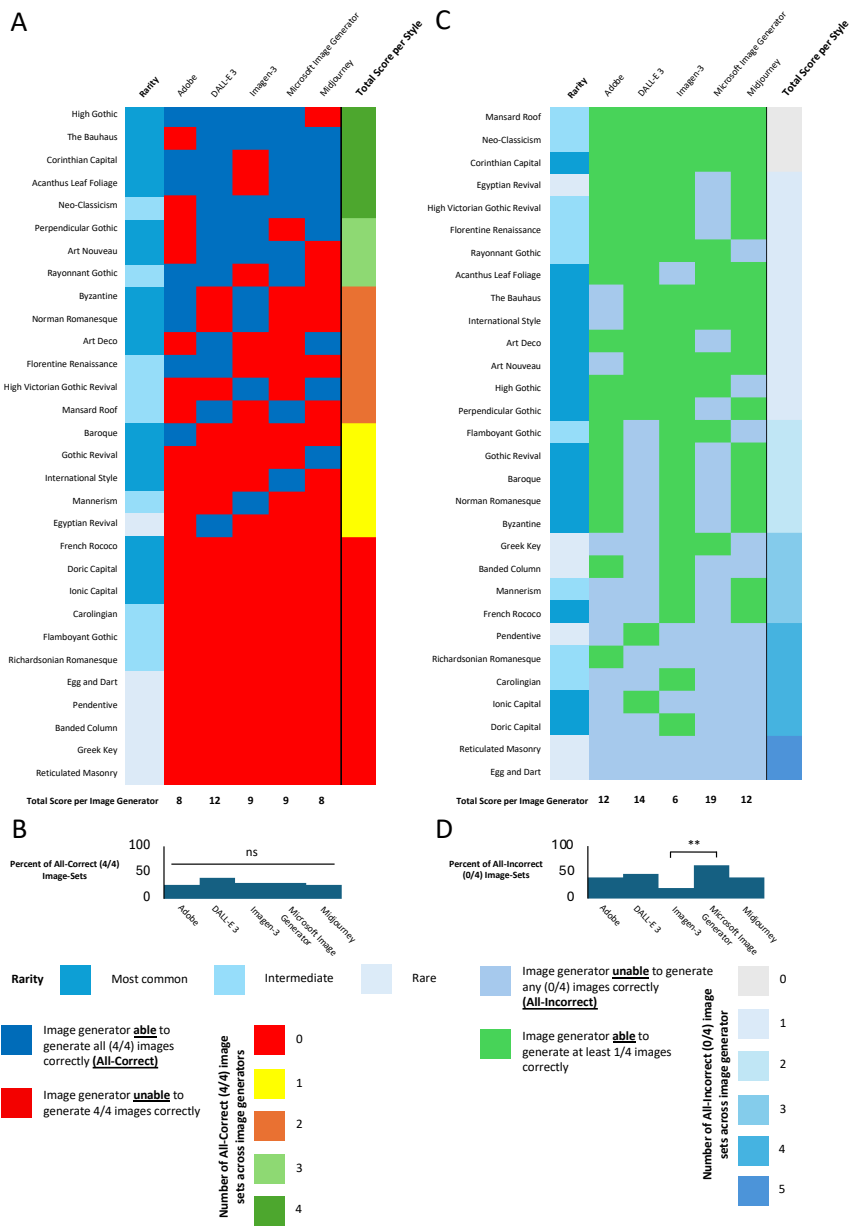

**Figure 6. Heatmaps of 'all-correct' (4/4) and 'all-incorrect' (0/4) outcomes by generator and prompt.** (A) Heatmap summarizes 'all-correct' outcomes (4/4 accurate images) for each prompt, with prompts labeled by rarity tier. (B) Percentage of prompts yielding 'all-correct' image sets (4/4) for each generator (n = 30 prompts per platform). (C) Heatmap summarizes 'all-incorrect' outcomes (0/4 accurate images) for each prompt, with prompts labeled by rarity tier. (D) Percentage of prompts yielding 'all-incorrect' image sets (0/4) for each generator (n = 30 prompts per platform). Pairwise differences in set-level accuracy between generators were assessed using paired t-tests, significance was reached at p < 0.05.

We also observed a correlation between 'all-correct' outcomes and our rarity tiering discussed earlier (Common, 31/75 (41%); Intermediate, 14/50 (28%); Rare, 1/25 (4%), p < 0.05), with significant differences between Common vs. Rare (p < 0.01) and Intermediate vs. Rare (p < 0.05) (**Table 2A**). We did observe a similar trend between 'all-incorrect' outcomes (Common, 26/75 (35%); Intermediate, 21/50 (42%); Rare, 16/25 (64%), p < 0.05), however individual comparisons did not reach significance when adjusting for multiple comparisons (**Table 2B**).

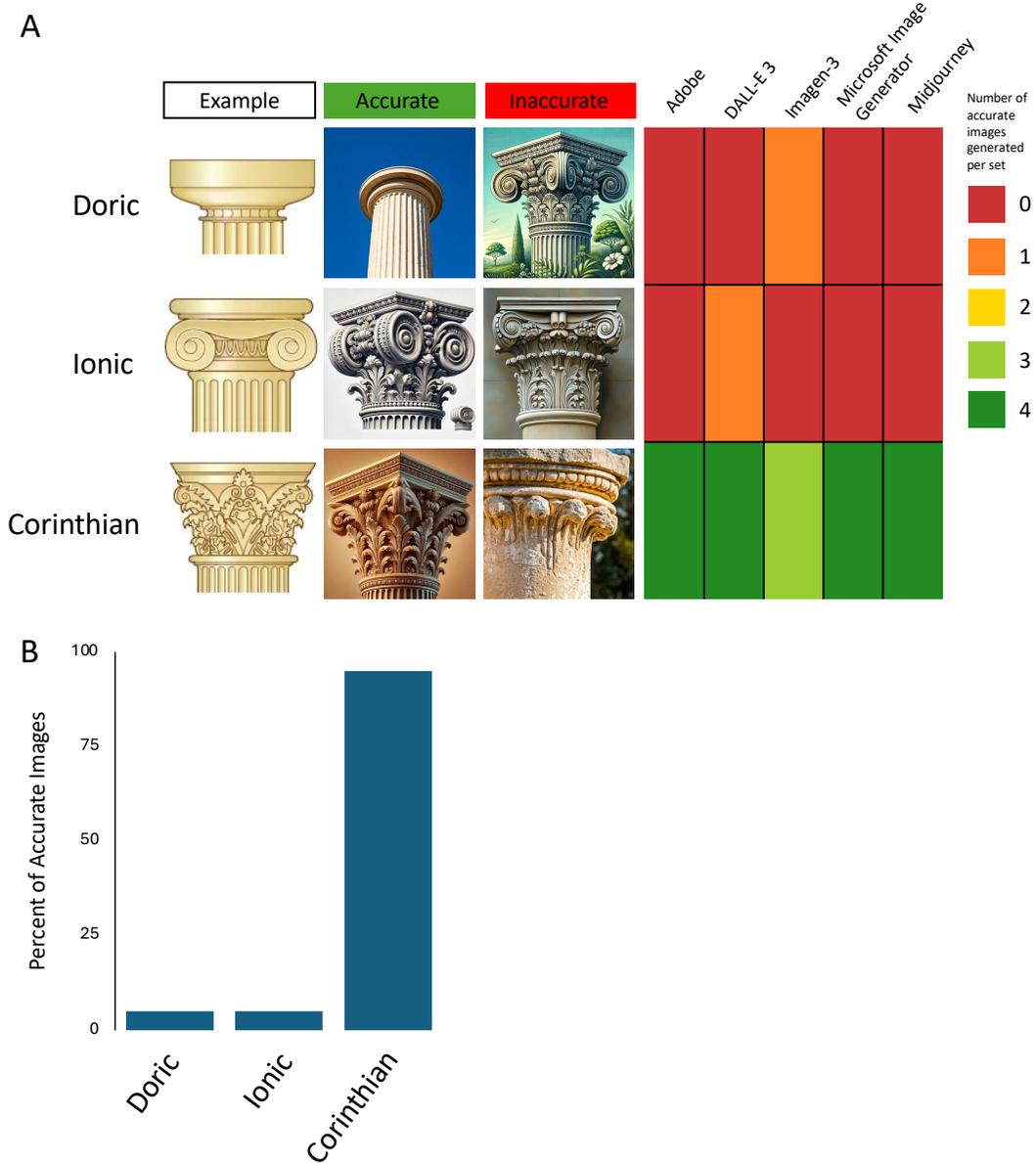

**Figure 7. Case Study: Doric, Ionic, & Corinthian Capitals across five GenAI text-to-image generators.** (A) Canonical reference schematics (left), representative "accurate" (centre) and "inaccurate" (right) generated examples of Doric, Ionic, & Corinthian Capitals. Heatmap (far right) reports set-level accuracy (0–4 accurate images per four-image set) for each platform on each order prompt. (B) Percent of accurate images by order across the full set of generated outputs for Doric, Ionic, & Corinthian Capitals.

> *Case Study: Doric, Ionic, & Corinthian Capitals*
>
> Our set of 30 prompts included Doric, Ionic, and Corinthian capitals, three of the most-recognizable and well-known architectural forms of the past two millennia (**Figure 7A**). All five text-to-image systems exhibited similar outcome frequencies in generating Doric, Ionic and Corinthian capitals; specifically, we recorded an overall success rate of 5% for Doric and Ionic capitals as compared to an overall success rate of 95% for Corinthian capitals across all five text-to-image systems (**Figure 7A-B**).
>
> This case-study may serve as a well-placed example of how systematic over-detailing and over-embellishment of ornamentation for simple architectural forms (e.g., Doric and Ionic capitals) can result in drastically inaccurate GenAI images, while generation of complex, embellished architectural forms (e.g., Corinthian capitals) can benefit from those same tendencies of GenAI text-to-image systems. A close examination of our Doric, Ionic, and Corinthian capital image sets revealed that inaccurate images were often outcomes where the GenAI text-to-image system seems to misinterpret the request for the more simple and historically earlier Doric capital and instead generates features of the more ornate Corinthian capital, such as a carved Acanthus leaf ornament and rounded volutes at the four compass points. This also raises questions about whether GenAI text-to-image systems are able to reliably generate simple and complex architectural forms, effectively discern between similar image prompts, and whether image training datasets were sufficiently well-annotated to allow GenAI systems to differentiate between visually similar architectural forms.

*Qualitative Observations*

A systematic visual review of our n = 600 image generation set yielded several interesting observations from a technical standpoint. When trying to reproduce historical architectural styles or carved ornament, we observed that the generated output is often inaccurate and over-embellished, sometimes almost to the point of caricature, as seen with the inaccurate examples in **Figure 2A**. However, the accurate Byzantine representations in **Figure 2A** do suggest that GenAI text-to-image systems are able to intermittently produce accurate façade decoration and pattern, while struggling to accurately represent the correct massing of volumes that follow a Byzantine plan; for the inaccurate example of a Byzantine architectural style, the over-elaborate decoration and massing illustrates the stochastic nature of GenAI text-to-image systems in over-exaggerating different abundant aspects (windows, arches, doors) and at times, auxiliary aspects (small decorative carvings, spires, minarets, sculptures) of the image.

Similarly, in its attempts to recreate historical architectural styles, GenAI text-to-image systems appear to rely heavily on the representation of carved motifs in and around doorways and windows. For instance, the accurate Norman Romanesque example in **Figure 3A** indicates that GenAI text-to-image systems are at times able to accurately represent three-dimensional space, creating multiple archivolts in the doorway supported by semi-circular pilasters, a rudimentary tympana above the door, and corbels that jut out of the façade in the fashion of early medieval buildings and including accurate shading and shadowing. Though these elements are also present in the inaccurate images of Norman Romanesque in **Extended Data 1**, again, we observed the frequent over-exaggeration of different abundant aspects of the image, such as windows, arches, and blind arches. A similar problem was also observed in the more modern architectural styles that emphasize floors, spare façades and windows; for example, several of the GenAI text-to-image systems generated inaccurate image-sets with additional floors for International Style buildings in **Extended Data 1**, seemingly modelled on Le Corbusier's Villa Savoye, a famous *two-story* structure.

Stylistic representations appear to follow a similar pattern as architectural forms and styles in our dataset. For example, in the accurate image of the Mansard Roof in **Figure 3A**, the basic structure is correct, however the outsized proportion and abundance of windows in and between the dormers appears somehow aesthetically 'off-putting'. Similarly, the accurate image of a Banded Column in **Figure 3A** does display raised bands, however they are unequally spaced at odd intervals, giving rise to a sort of 'uncanny' or 'otherworldly' aspect to the image. By contrast, both inaccurate images seem to both misinterpret and over-embellish the prompts, with too many windows for the Mansard Roof image, and an overly colorful, almost carnival-like looping design scheme and lozenge-like pattern on the column having little in common with reserved neo-classical geometry for the Banded Column.

Additionally, we observed in **Extended Data 1** a tendency for GenAI text-to-image systems to confuse aspects of medieval architecture and their later (often 19$^{th}$ century) revivals, which are quite different aesthetically. For example, the inaccurate Gothic Revival example in **Figure 3A** appears more historically connected to the Flamboyant Gothic medieval style of the late 15$^{th}$ century, complete with skeletal wall technology allowing for large windows, pointed finials, and proliferation of stone tracery. This is not to say that some of these elements do not appear in Gothic Revival churches of the 19$^{th}$ century, but they tend not to appear in moderation in the GenAI images in our dataset.

## Discussion

This study's quantitative and qualitative comparative approach, utilizing 600 GenAI images across five GenAI text-to-image systems and 30 prompts, offers a detailed look into the strengths, weaknesses and limitations of GenAI text-to-image systems, and allows us to move beyond the concept of GenAI 'hallucinations' as a catch-all term for GenAI inaccuracies. Rather, we posit that, particularly for highly precise, detail-oriented GenAI images as in our architectural style and form dataset, GenAI text-to-image systems generate images as if *inside a mirage*, that is to say, GenAI text-to-image systems are not quite able to perceive the true form of the image they attempt to copy, and therefore reproduce some of the essential qualities of the image with, at times significant, distortion or over-embellishment.

This study highlights several key trends in GenAI text-to-image generation as it pertains to architectural images of style and form. First, within our groupings by Rarity, prompts grouped as Rare (e.g., Greek key, banded column, pendentive) had lower per-style means than prompts grouped as Common, were significantly less likely to achieve an 'all-correct' (4/4) image-set, and seemed to yield more extreme examples of failure in our qualitative review. This pattern aligns with our qualitative observations that GenAI text-to-image systems gravitate toward over-embellishment and struggle with rarer and/or less intuitively-named, but more descriptive, architectural forms, such as 'Egg-and-Dart' or 'Banded Column'. This finding raises concerns that GenAI text-to-image systems may struggle with reproducing rarer, less well-known, or esoteric styles, forms and motifs in the architectural cannon.

Second, in our study no GenAI platform was able to achieve an accuracy score above 52%, the aggregate average of all five GenAI text-to-image systems was closer to 42%, and the aggregate average for 'all-correct' (4/4) image-sets was less than 31% across all five GenAI text-to-image systems. While our sample size was small, this finding should prompt users of GenAI text-to-image systems, educators, and students to take great care when considering the use of GenAI text-to-image systems to generate images from the architectural cannon. A well-studied phenomena in computer vision may help explain these first two observations; because modern computer vision exhibits a bias toward local textures over global shapes, this can reward decorative elaboration even as it erodes GenAI text-to-image system fidelity to plan, section, and proportion out images[31,33,37].

Third, while our selected GenAI text-to-image systems produced similar frequencies of 'all-correct' (4/4) image-set outcomes, they appeared to differ more in frequency of 'all-incorrect' (0/4) image-set outcomes, which appeared to drive differences in overall success rates for platforms (with Imagen 3 the most-reliable, and Microsoft Image Generator the least-reliable). Indeed, we observed that for many platforms, particularly Microsoft Image Generator and DALL-E 3, accuracy of image generation appeared to be an 'all-or-nothing' event – that is to say, these individual GenAI text-to-image systems frequently either succeeded for all 4/4 images in the set (31%) or catastrophically misrepresented the prompt across the board, failing for all images 0/4 in the set (42%), with only a minority scoring in-between (27%).

One possible explanation for this phenomenon is 'distributional inertia' in the learned generative prior: when a prompt lies in a well-represented region of the training distribution, sampling is strong and stable, yielding consistently aligned outputs, whereas when the prompt targets a rarer or weakly covered concept, the model tends to revert to default prototypes, producing a repeated set of often widely divergent and misaligned images, rather than a graded spectrum of partially correct and partially incorrect images in a single set – hence, an 'all or nothing' outcome[38,39]. Differences in training datasets between individual GenAI text-to-image systems may therefore account for the differences in reliability between platforms that we observed in this study.

A deeper dive into our n = 600 image generation set (**Extended Data 1**) supports this concept of 'distributional inertia' to some extent; a systematic visual review of all 600 images suggested that higher-accuracy prompts (e.g., High Gothic, The Bauhaus) may produce less intra-image-set heterogeneity than lower-accuracy prompts (e.g., Reticulated Masonry, Carolingian), similar to results observed in Jiang *et al.*, (2025) and Wenger & Kenett (2025) in text-based generative large language models (LLMs)[40,41]. Similar to LLM 'hallucinations', we observed that in some cases, GenAI text-to-image systems appeared to become 'entrenched' into incorrect interpretations of the prompt with a high degree of intra-image-set homogeneity (e.g., Egg and Dart [Adobe Firefly], Ionic Capital [Microsoft Image Generator]). Difficulties in generating accurate Doric and Ionic capitals provides another example of this 'entrenchment' phenomenon of GenAI text-to-image systems: a large proportion of the Doric and Ionic images generated (~60% of Doric and Ionic capitals) instead resembled Corinthian-like capitals, mostly stemming from individual image-sets with a high degree of intra-image-set homogeneity (e.g., Doric Capital [Midjourney], Ionic Capital [Adobe Firefly]).

Taken together, our results suggest that current GenAI text-to-image systems are *unreliable partners* for image generation, particularly when the target image is from a style or form with a clear rule-bound structure, as with architectural imagery of buildings, styles, forms, and motifs.

## Limitations

This study has several limitations. We evaluated 30 prompts, weighted toward Euro-American styles and canonical motifs. The results should not be over-generalized to under-represented architectural traditions, particularly non-Western styles, or to non-architectural domains. Future studies could expand the prompt generation set with a wider variety of architectural styles, a larger set of prompts (beyond n = 30), or an increase in the number of sampled outputs per prompt (beyond n = 4) to reduce variability and better characterize variance among GenAI text-to-image systems. Future research could also explicitly compare archived model versions over time (e.g., ChatGPT 4.0 vs 5.0), incorporate controlled prompt-iteration protocols to allow for model iteration after first-failed attempts, and repeat the evaluation longitudinally over a period of months or years to quantify whether GenAI text-to-image fidelity improves across successive model releases. We reported only consensus scores from two expert architectural historians as part of our ratings process (see **Methods**). While we elected to employ this method to strengthen robustness of conclusions and internal consistency of ratings, future studies may wish to utilize a more formal ratings rubric or a wider pool of raters with an inter-rater statistic to quantify reliability more precisely, rather than a consensus-based ratings system. Further, automated alignment metrics (e.g., VQAScore, TIFA) were not applied to our dataset; future work could integrate expert ratings with such metrics to further assess fidelity and reliability.

## Conclusions

Because architectural knowledge is image-borne, that is, shared, taught, and retrieved through pictures, unlabeled and labeled synthetic images pose challenges for pedagogy in this field. Educators should take care whenever using GenAI images or image-to-text generators to treat such images as interpretations, rather than examples, integrate provenance literature and training (e.g., how to read and verify Content Credentials/SynthID) into survey and studio courses, and to consider broaching the subject of GenAI images with paired exemplars (accurate images vs. near-miss images vs. incorrect images) to bulwark student understanding. When coupled with the uneven expansion of visible and invisible labels for GenAI images (e.g., C2PA Content Credentials, SynthID watermarks, and platform-specific labels), our results argue for style-aware benchmarks as a guardrail for image use in pedagogy and public scholarship. Minimal-accuracy tests, such as 'does the Romanesque arch resolve correctly in this image?' or 'does a Doric echinus remain plain in this image?' are easy for experts to check, and hard for models to fake repeatedly. Such tests complement already extant, but limited, automated metrics such as VQAScore and GenAI-Bench, T2I-CompBench++, and TIFA that quantify text–image alignment and compositional fidelity, but do not capture historically grounded rule compliance without expert input[3,42-46]. This is in line with UNESCO's 2023 Guidance for Generative AI in Education and Research, which likewise urges human-centered adoption and curricular adjustments to protect validation of knowledge[47].

The reliability concerns and issues for GenAI images highlighted by this study are not limited to the fields of architecture and architectural history, and are likely to arise in any field that depends on precise, rule-bound visual representation. As a consequence, there will likely be increased reliance on textbooks and other non-online or semi-offline forms of knowledge recording and transmission; however, this shift may exacerbate accessibility and equity concerns by reintroducing higher financial, geographic, and institutional barriers to access[48,49]. Comparable reliability problems should be anticipated in other detail-oriented professions, such as archaeology, art and museum history and conservation studies, geology, taxonomy, forensic science, and particularly medicine, where fine-grained visual and morphological distinctions are a critical component of diagnostic criteria. In internal medicine and infectious diseases, clinical reasoning frequently depends on subtle, high-stakes pattern recognition (e.g., lesion morphology, distributional patterns, and imaging correlates), and recent safety-oriented reviews emphasize that generative systems can be inconsistent and error-prone in ways that directly implicate quality of care and patient safety if they are treated as authoritative or are introduced into workflows without robust validation[50-52]. Dermatology is an especially clear analogue to our findings because dermatological diagnostic work is extremely visually-oriented; in an experimental study of AI-generated dermatologic conditions, researchers reported both deficient skin-tone diversity and poor diagnostic identifiability of intended conditions, while related methodological work shows that bias in GenAI medical imagery can systematically underrepresent darker skin tones, limitations which would be unacceptable if such images were used for clinician training, patient education, or downstream model development without careful curation and audit[53]. As health-themed GenAI images become more commonplace across media, educational materials, and even academic contexts, our results support emerging guidance on synthetic content emphasizing provenance, labeling, and dataset governance as key prerequisites for maintaining professional standards and limiting harm as GenAI imagery increasingly circulates online or is repurposed for training data for future GenAI text-to-image systems.

**Table 1.** Prompt-generation pairs (n=30) for our image generation dataset.

| Architectural Styles | Descriptive Scoring Criteria | Prompt Stem |
|---|---|---|
| Byzantine | Domed roofs, symmetrical planning; Greek cross plan of square central mass with four arms of equal length, solid, soaring spaces, defense-like massing, sumptuous decoration that included marble columns and inlay, mosaics on the vaults, inlaid-stone pavements and gold coffered ceilings. | Create an image of a building in the style of [architectural style] |
| Norman Romanesque | Round-headed arches in surging rhythm, repeated arches, barrel-vaulted ceiling, multi-ordered entrances, longitudinal planning; rectangular nave with short transept arms, carved ornament; geometric, animal and imagined/mythic creatures. | Create an image of a building in the style of [architectural style] |
| Carolingian | Early 4th century architecture associated with Constantine's Christian Empire, Germanic in nature, featuring a revival of Roman motifs – round arches articulated with patterned and coloured stonework and marble revetment, in a bold, no-nonsense manner. | Create an image of a building in the style of [architectural style] |
| Perpendicular Gothic | The final phase of English Gothic architecture, featuring fine intricate stonework window tracery, elaborate fan vaults, and interior wall tracery. | Create an image of a building in the style of [architectural style] |
| Rayonnant Gothic | Phase of French Gothic architecture noted for thin wall technology and soaring height, featuring pointed window dressing. | Create an image of a building in the style of [architectural style] |
| High Gothic | Phase of English Gothic architecture noted for introducing thin wall technology, featuring increasingly elaborate geometrical decoration to the structural forms that had been established during the preceding century. | Create an image of a building in the style of [architectural style] |
| Flamboyant Gothic | Late phase of Gothic architecture dominated by emphasis on decoration, especially stone window tracery in flamelike S-shaped curves. | Create an image of a building in the style of [architectural style] |
| Florentine Renaissance | Urban Italian city-state movement associated with emerging wealth amidst political struggles and use of public funds in religious architecture featuring a re-birth of Roman ideals of symmetry, proportion, and geometry. | Create an image of a building in the style of [architectural style] |
| Mannerism | Architectural style featuring complexity and novelty, challenging the rules of Renaissance proportion. | Create an image of a building in the style of [architectural style] |
| Baroque | Architectural style featuring vaulted cupolas, twirling and swirling colonnades on undulating facades, and ornate frescoes and painted ceilings in trompe l'oeil. | Create an image of a building in the style of [architectural style] |
| French Rococo | Delicate interlacing of intricate curves and counter-curve decoration, shell forms and natural shapes, and asymmetrical design. | Create an image of a building in the style of [architectural style] |
| Neo-Classicism | Symmetrical planning and placement of geometric forms often on a grand scale, return of Classical motifs – orders, and dramatic use of ornament. | Create an image of a building in the style of [architectural style] |
| Gothic Revival | Architectural style featuring pointed arches, vaults, asymmetrical planning, and structural ornament – buttresses, rib vaults, and window decoration based on surviving medieval buildings. | Create an image of a building in the style of [architectural style] |
| Egyptian Revival | Architectural style featuring smooth, monolithic surfaces, monumentality, and ornament referencing Ancient Egyptian, Roman and Greek motifs. | Create an image of a building in the style of [architectural style] |
| High Victorian Gothic Revival | Heavier and more substantial than early Gothic Revival, featuring polychromatic stonework colour schemes, decorative dormers and gables, round turrets with conical roofs, and corbelled brickwork. | Create an image of a building in the style of [architectural style] |
| Art Nouveau | Architectural style featuring sinuous, sculptural, organic shapes, arches, curving lines, and sensual ornamentation. | Create an image of a building in the style of [architectural style] |
| Art Deco | Simple, clean shapes, often with a "streamlined" look, simplification of form and space into geometrical language, and slightly curved corners, thinly articulated linear and geometric patterns. | Create an image of a building in the style of [architectural style] |
| Richardsonian Romanesque | Round-headed arches often with thick stubby columns, asymmetrical planning, rough or hammer-dressed stonework, and predominance of sandstone material for carved ornament mimicking Romanesque motifs. | Create an image of a building in the style of [architectural style] |
| International Style | Rectilinear forms, light, taut plane surfaces completely stripped of applied ornamentation, open interior spaces and visually weightless quality from cantilevered construction. | Create an image of a building in the style of [architectural style] |
| The Bauhaus | Originated with a school of art and architecture in Germany in 1919 that developed curriculum devoted to improving everyday design of objects, font, and construction comprised of sleek lines, experimental use of materials, disruption to traditional standards and methods of supporting walls such as curtain walls and hollowed out corners. | Create an image of a building in the style of [architectural style] |
| **Architectural Motifs** | **Descriptive Scoring Criteria** | **Prompt Stem** |
| Mansard Roof | Roof with two slopes on every side, the lower slope being considerably steeper than the upper (sometimes pierced by a dormer window), differs from the gambrel by displaying the same profile on all sides. | Create an image of a [architectural motif] |
| Doric Capital | The uppermost member of a column supporting the architrave, comprised of a round Echinus tapering upward surmounting a square Abacus. | Create an image of a [architectural motif] |
| Ionic Capital | The uppermost member of a column supporting the architrave, comprised of a large double scroll or volute, between the Echinus and the Abacus, which projects strongly beyond the width of the shaft or column. | Create an image of a [architectural motif] |
| Corinthian Capital | The uppermost member of a column supporting the architrave, comprised of an inverted bell covered with the curly shoots and leaves of the acanthus plant, which seem to sprout from the top of the column shaft. | Create an image of a [architectural motif] |
| Greek Key | Consisting of lengths of straight lines or narrow bands, usually connected and at right angles to each other in T, L, or square-cornered G shapes, so arranged that the spaces between the lines or bands are approximately equal to the width of the bands. | Create an image of a [architectural motif] |
| Egg and Dart | Consists of a series of bas-relief ovals alternating with pointed, narrow, dartlike carvings. | Create an image of a [architectural motif] |
| Pendentive | A means of structurally supporting a transition from a 90 degree corner to a round dome by filling the upper corner of a room with a form that tapers downward. | Create an image of a [architectural motif] |
| Banded Column | A smooth shaft or pilaster (supportive member) comprised of alternating bands (or rings) that project beyond the surface area of the main element. | Create an image of a [architectural motif] |
| Reticulated Masonry | The carving or finishing of building stones with irregular grooves intended to resemble worm tracks. | Create an image of a [architectural motif] |
| Acanthus Leaf Foliage | Decorative plant-based ornament with stylized characteristics of Mediterranean foliage, having jagged leaves. | Create an image of a [architectural motif] |

**Table 2.** Statistical analysis of 'all-correct' (4/4) and 'all-incorrect' (0/4) outcomes stratified by rarity tiering (Common, Intermediate, Rare). Differences in the proportion of (A) 'all-correct' (4/4) and (B) 'all-incorrect' (0/4) image sets across rarity tiers were evaluated using a chi-square test of independence, and an ordered chi-square test for trend (Cochran–Armitage) was performed to assess monotonic changes across tiers. Pairwise comparisons between tiers were conducted using two-sided Fisher's exact tests, with Holm correction applied to control the family-wise error rate, significance was reached at p < 0.05.

| Table 2A: 'All-correct' outcomes | | | |
|---|---|---|---|
| Tier | All-correct' (4/4) outcome | Any other outcome | Proportion of 'All-correct' (4/4) outcomes |
| Common | 31 | 44 | 0.413 (41%) |
| Intermediate | 14 | 36 | 0.280 (28%) |
| Rare | 1 | 24 | 0.040 (4%) |
| Test | Statistic | p-value | |
| Chi-square test of independence | $\chi^2$ = 12.542 | 0.00189 | |
| Chi-square test for trend (Cochran–Armitage) | $\chi^2$ = 12.140 | 0.000493 | |
| Comparison | All-correct' (4/4) outcome rate | Odds ratio | p-adjust value |
| Common vs Intermediate | 41% vs 28% | 1.812 | 0.18277 |
| Common vs Rare | 41% vs 4% | 16.909 | **0.00105** |
| Intermediate vs Rare | 28% vs 4% | 9.333 | **0.02961** |

| Table 2B: 'All-incorrect' outcomes | | | |
|---|---|---|---|
| Tier | All-incorrect' (0/4) outcome | Any other outcome | Proportion of 'All-incorrect' (0/4) outcomes |
| Common | 26 | 49 | 0.347 (35%) |
| Intermediate | 21 | 29 | 0.420 (42%) |
| Rare | 16 | 9 | 0.640 (64%) |
| Test | Statistic | p-value | |
| Chi-square test of independence | $\chi^2$ = 6.623 | 0.03646 | |
| Chi-square test for trend (Cochran–Armitage) | $\chi^2$ = 5.921 | 0.01496 | |
| Comparison | All-incorrect' (0/4) outcome rate | Odds ratio | p-adjust value |
| Common vs Intermediate | 35% vs 42% | 0.733 | 0.45364 |
| Common vs Rare | 35% vs 64% | 0.298 | 0.05451 |
| Intermediate vs Rare | 42% vs 64% | 0.407 | 0.17925 |

**Extended Data 1.** Full (n = 600) image generation dataset. For access, please contact the lead author at jamie.magrill@mail.mcgill.ca

# References

1. Rombach R, Blattmann A, Lorenz D, Esser P, Ommer B. High-Resolution Image Synthesis with Latent Diffusion Models. 2021:arXiv:2112.10752. doi:10.48550/arXiv.2112.10752 Accessed December 01, 2021. https://ui.adsabs.harvard.edu/abs/2021arXiv211210752R
2. Rombach R, Blattmann A, Lorenz D, Esser P, Ommer B. High-resolution image synthesis with latent diffusion models. 2022:10684-10695.
3. Frolov S, Hinz T, Raue F, Hees J, Dengel A. Adversarial text-to-image synthesis: A review. *Neural Networks*. 2021/12/01/ 2021;144:187-209. doi:https://doi.org/10.1016/j.neunet.2021.07.019
4. Yang L, Zhang Z, Song Y, et al. Diffusion Models: A Comprehensive Survey of Methods and Applications. *ACM Comput Surv*. 2023;56(4):Article 105. doi:10.1145/3626235
5. Sriram A. Google to pause Gemini AI model's image generation of people due to inaccuracies. *Reuters*. https://www.reuters.com/technology/google-pause-gemini-ai-models-image-generation-people-2024-02-22/
6. Raghavan P. Gemini image generation got it wrong. We'll do better. 2024. https://blog.google/products/gemini/gemini-image-generation-issue/
7. Kostof S. *A History of Architecture: Settings and Rituals*. 2 ed. Oxford University Press; 1995.
8. Trachtenberg M, Hyman I. *Architecture, from prehistory to post-modernism : the Western tradition*. 2 ed. Harry N. Abrams; 1986.
9. Fletcher B. *Banister Fletcher's A History of Architecture*. B. T. Batsford Ltd; 1943.
10. Viollet-le-Duc E-E. *Dictionnaire raisonné de l'architecture française du XIe au XVIe siècle*. A. Morel; 1875.
11. Frankl P. *Gothic Architecture*. Penguin Books; 1962.
12. Summerson J. *The Classical Language of Architecture*. Thames & Hudson; 1963.
13. Jones O. *The Grammar of Ornament*. Day and Son; 1856.
14. Frampton K. *Studies in Tectonic Culture: The Poetics of Construction in Nineteenth and Twentieth Century Architecture*. MIT Press; 1995.
15. Magrill B. *A Commerce of Taste: Church Architecture in Canada, 1867-1914*. McGill-Queen's University Press; 2012.
16. Pollio V. *The Ten Books on Architecture*. Morgan MH. Harvard University Press; 1914.
17. Palladio A. *The four books of architecture*. Dover Publications; 1965.
18. Wölfflin H. *Classic Art: An Introduction to the Italian Renaissance*. Murray P, Murray L. 2 ed. Phaidon Press; 1952.
19. Gombrich EH. *The Story of Art*. 16 ed. Phaidon Press; 1995.
20. Wittkower R. *Architectural Principles in the Age of Humanism*. The Warburg Institute, University of London; 1949.
21. Janson HW, Janson AF. *History of Art*. 4 ed. Abrams; 1991.
22. Panofsky E. *Gothic Architecture and Scholasticism*. The Archabbey Press; 1951.
23. Stokstad M, Cothren MW. *Art History*. 6 ed. Pearson; 2018.
24. Regulation (EU) 2024/1689 of the European Parliament and of the Council of 13 June 2024 laying down harmonised rules on artificial intelligence (Artificial Intelligence Act) (Official Journal of the European Union) (2024).

25. Executive Order 14110 of October 30, 2023, on Safe, Secure, and Trustworthy Development and Use of Artificial Intelligence (2023).
26. Highlights of the 2023 Executive Order on Artificial Intelligence for Congress (2024).
27. Blueprint for an AI Bill of Rights: Making Automated Systems Work for the American People. Washington, DC: White House; 2022.
28. Artificial Intelligence Risk Management Framework (NIST AI 100-1) (National Institute of Standards and Technology) (2023).
29. Advancing Governance, Innovation, and Risk Management for Agency Use of Artificial Intelligence (Office of Management and Budget) (2024).
30. Rijsbosch B, van Dijck G, Kollnig K. Adoption of Watermarking for Generative AI Systems in Practice and Implications under the new EU AI Act. 2025:arXiv:2503.18156. doi:10.48550/arXiv.2503.18156 Accessed March 01, 2025. https://ui.adsabs.harvard.edu/abs/2025arXiv250318156R
31. Geirhos R, Rubisch P, Michaelis C, Bethge M, Wichmann FA, Brendel W. ImageNet-trained CNNs are biased towards texture; increasing shape bias improves accuracy and robustness. 2018:arXiv:1811.12231. doi:10.48550/arXiv.1811.12231 Accessed November 01, 2018. https://ui.adsabs.harvard.edu/abs/2018arXiv181112231G
32. Malhotra G, Evans BD, Bowers JS. Hiding a plane with a pixel: examining shape-bias in CNNs and the benefit of building in biological constraints. *Vision Res*. Sep 2020;174:57-68. doi:10.1016/j.visres.2020.04.013
33. Baker N, Lu H, Erlikhman G, Kellman PJ. Deep convolutional networks do not classify based on global object shape. *PLoS Comput Biol*. Dec 2018;14(12):e1006613. doi:10.1371/journal.pcbi.1006613
34. Taeihagh A. Governance of Generative AI. *Policy and Society*. 2025;44(1):1-22. doi:10.1093/polsoc/puaf001
35. Shumailov I, Shumaylov Z, Zhao Y, Papernot N, Anderson R, Gal Y. AI models collapse when trained on recursively generated data. *Nature*. Jul 2024;631(8022):755-759. doi:10.1038/s41586-024-07566-y
36. Shumailov I, Shumaylov Z, Zhao Y, Gal Y, Papernot N, Anderson R. The Curse of Recursion: Training on Generated Data Makes Models Forget. 2023:arXiv:2305.17493. doi:10.48550/arXiv.2305.17493 Accessed May 01, 2023. https://ui.adsabs.harvard.edu/abs/2023arXiv230517493S
37. Brendel W, Bethge M. Approximating CNNs with Bag-of-local-Features models works surprisingly well on ImageNet. 2019:arXiv:1904.00760. doi:10.48550/arXiv.1904.00760 Accessed March 01, 2019. https://ui.adsabs.harvard.edu/abs/2019arXiv190400760B
38. Creswell A, White T, Dumoulin V, Arulkumaran K, Sengupta B, Bharath AA. Generative Adversarial Networks: An Overview. *IEEE Signal Processing Magazine*. 2018;35(1):53-65. doi:10.1109/MSP.2017.2765202
39. Fu S, He X, Hu H. LTB-Solver: Long-tailed Bias Solver for image synthesis of diffusion models. *Neurocomputing*. 2025/06/14/ 2025;634:129651. doi:https://doi.org/10.1016/j.neucom.2025.129651
40. Jiang L, Chai Y, Li M, et al. Artificial Hivemind: The Open-Ended Homogeneity of Language Models (and Beyond). 2025:arXiv:2510.22954. doi:10.48550/arXiv.2510.22954 Accessed October 01, 2025. https://ui.adsabs.harvard.edu/abs/2025arXiv251022954J

41. Wenger E, Kenett Y. We're Different, We're the Same: Creative Homogeneity Across LLMs. 2025:arXiv:2501.19361. doi:10.48550/arXiv.2501.19361 Accessed January 01, 2025. https://ui.adsabs.harvard.edu/abs/2025arXiv250119361W

42. Hartwig S, Engel D, Sick L, et al. A Survey on Quality Metrics for Text-to-Image Generation. *IEEE Trans Vis Comput Graph*. Oct 2025;31(10):9464-9483. doi:10.1109/tvcg.2025.3585077

43. Huang K, Duan C, Sun K, Xie E, Li Z, Liu X. T2I-CompBench++: An Enhanced and Comprehensive Benchmark for Compositional Text-to-Image Generation. *IEEE Trans Pattern Anal Mach Intell*. May 2025;47(5):3563-3579. doi:10.1109/tpami.2025.3531907

44. Sukkar AW, Fareed MW, Yahia MW, Abdalla SB, Ibrahim I, Senjab KAK. Analytical Evaluation of Midjourney Architectural Virtual Lab: Defining Major Current Limits in AI-Generated Representations of Islamic Architectural Heritage. *Buildings*. 2024;14(3):786.

45. Lin Z, Pathak D, Li B, et al. Evaluating Text-to-Visual Generation with Image-to-Text Generation. 2024:arXiv:2404.01291. doi:10.48550/arXiv.2404.01291 Accessed April 01, 2024. https://ui.adsabs.harvard.edu/abs/2024arXiv240401291L

46. Hu Y, Liu B, Kasai J, et al. TIFA: Accurate and Interpretable Text-to-Image Faithfulness Evaluation with Question Answering. 2023:arXiv:2303.11897. doi:10.48550/arXiv.2303.11897 Accessed March 01, 2023. https://ui.adsabs.harvard.edu/abs/2023arXiv230311897H

47. UNESCO. *Guidance for generative AI in education and research*. 2023. https://www.unesco.org/en/articles/guidance-generative-ai-education-and-research

48. Hilton J. Open educational resources and college textbook choices: a review of research on efficacy and perceptions. *Educational Technology Research and Development*. 2016/08/01 2016;64(4):573-590. doi:10.1007/s11423-016-9434-9

49. Burns T, Gottschalk F, eds. *Education in the Digital Age: Healthy and Happy Children*. OECD Publishing; 2020.

50. Scott IA, Reddy S, Kelly T, Miller T, van der Vegt A. Using generative artificial intelligence in clinical practice: a narrative review and proposed agenda for implementation. *Med J Aust*. Dec 8 2025;223(11):664-672. doi:10.5694/mja2.70057

51. Howell MD. Generative artificial intelligence, patient safety and healthcare quality: a review. *BMJ Quality & Safety*. 2024;33(11):748-754. doi:10.1136/bmjqs-2023-016690

52. Morgan DJ, Goodman KE, Branch-Elliman W, et al. Using Generative Artificial Intelligence for Healthcare-Associated Infection Surveillance. *Clinical Infectious Diseases*. 2025;doi:10.1093/cid/ciaf619

53. Joerg L, Kabakova M, Wang JY, et al. AI-generated dermatologic images show deficient skin tone diversity and poor diagnostic accuracy: An experimental study. *J Eur Acad Dermatol Venereol*. Dec 2025;39(12):2134-2141. doi:10.1111/jdv.20849